\documentclass[acmsmall]{acmart}
\usepackage{bm}
\usepackage{graphicx}
\usepackage{tabularx}
\usepackage{multirow}

\usepackage{subfigure}

\usepackage{algorithm}
\usepackage{algorithmic}

\usepackage{color}

\AtBeginDocument{%
  \providecommand\BibTeX{{%
    \normalfont B\kern-0.5em{\scshape i\kern-0.25em b}\kern-0.8em\TeX}}}

 \acmYear{2020}

\begin{document}


\title{RGCF: Refined Graph Convolution Collaborative Filering with Concise and Expressive Embedding}


\author{Kang Liu}
\affiliation{%
  \institution{Hefei University of Technology}
  \department{School of Computer and Information}
  \streetaddress{420 Feicui Road}
  \city{Hefei}
  \country{China}}
  \email{2018110917@mail.hfut.edu.cn}
  \postcode{230009}

\author{Feng Xue }
\affiliation{%
  \institution{Hefei University of Technology}
  \department{School of Computer and Information}
  \streetaddress{420 Feicui Road}
  \city{Hefei}
  \country{China}}
\email{feng.xue@hfut.edu.cn}
\postcode{230009}

\author{Richang Hong}
\affiliation{%
  \institution{Hefei University of Technology}
  \department{School of Computer and Information}
  \streetaddress{420 Feicui Road}
  \city{Hefei}
  \country{China}}
\email{hongrc@hfut.edu.cn}
\postcode{230009}

\thanks{The corresponding author is Feng Xue}
\thanks{this work is supported by the National Natural Science Foundation of China (Grants No. 61772170)}

\renewcommand{\shortauthors}{Kang Liu, Feng Xue, and Richang Hong.}

\begin{abstract}
	Graph Convolution Network (GCN) has attracted significant attention and become the most popular method for learning graph representations. In recent years, many efforts have been focused on integrating GCN into the recommender tasks and have made remarkable progress. At its core is to explicitly capture high-order connectivities between the nodes in user-item bipartite graph. However, we theoretically and empirically find an inherent drawback existed in these GCN-based recommendation methods, where GCN is directly applied to aggregate neighboring nodes will introduce noise and information redundancy. Consequently, these models' capability of capturing high-order connectivities among different nodes is limited,  leading to suboptimal performance of the recommender tasks. The main reason is that the the nonlinear network layer inside GCN structure is not suitable for extracting non-sematic features(such as one-hot ID feature) in the collaborative filtering scenarios. 

    In this work, we develop a new GCN-based Collaborative Filtering model, named \textbf{R}efined \textbf{G}raph convolution \textbf{C}ollaborative \textbf{F}iltering(RGCF), where the construction of the embeddings of users (items) are delicately redesigned from several aspects during the aggregation on the graph. Compared to the state-of-the-art GCN-based recommendation, RGCF is more capable for capturing the implicit high-order connectivities inside the graph and the resultant vector representations are more expressive. We conduct extensive experiments on three public million-size datasets, demonstrating that our RGCF significantly outperforms state-of-the-art models. We release our code at https://github.com/hfutmars/RGCF.
\end{abstract}




\maketitle

\section{Introduction}
Modern recommendation system has been widely applied to many online services, such as video recommendation \cite{youtube}, music recommendation \cite{music}, E-commerce \cite{visual2}, and social network \cite{social}. Collaborative Filtering (CF) is the mainstream of modern recommendation algorithms \cite{cf}\cite{scf}. The basic assumption of CF is that similar users would exhibit similar interest on same items. Matrix Factorization (MF) is the most classical CF method, which vectorizes all users and items only with their ID features, and reconstruct their historical interactions with the inner product of them \cite{mf}. MF can achieve a good performance with sufficient interaction data. However, since the issue of sparsity is ubiquitous in modern recommendations, MF fails to learn expressive vector representations for users and items.

\subsection{Why Graph Convolution Networks?}
In order to solve the performance bottleneck caused by sparsity of datasets, many efforts have been devoted to constructing complex embedding functions. Specifically, integrating all available useful information into the embedding representations can improve the model performance. SVD++ \cite{svd++} is the pioneer work that incorporates user historically interacted items into user's embedding construction to model his/her preference to get expressive embedding. However, SVD++ only encodes explicit connectivities between user and item into the embedding function, while forgoing the modeling of the implicit connectivities, which can be viewed as the paths between current node and its multi-hop neighboring nodes in user-item bipartite graph (aka. High-order connectivities). Graph-based methods \cite{birank}\cite{hoprec} are capable of capturing such high-order connectivities due to its capability of learning path information. For example, HOP-Rec \cite{hoprec} indirectly integrates high-order connectivities into the embedding learning process by using random walk to enrich the interaction data for a user with multi-hop connected items. 
Apart from Graph-based methods which indirectly using high-order connectivities to enrich training data, GCN-based methods \cite{gcmc}\cite{pinsage}\cite{gcn} directly encode high-order connectivities into the embeddings function and achieve the significant improvement against other CF methods, illustrating that GCN is the state-of-the-art approach for capturing high-order connectivities inside the user-item interaction graph structure.

\color{black}
\subsection{Why Not the Nonlinear Network Layers of GCN?}
It is worth mentioning that in some GCN-based machine learning tasks, such as image classification \cite{image-c} and node classification \cite{gcn}, nonlinear network layers is necessary for feature extraction since the initial vector representations contain abundant and diverse information.
In contrast, IDs of users(or items) used by most CF methods carry no complicated patterns or diverse semantic information that can be mined. 
We argue that directly using nonlinear graph convolution layers to process ID features like \cite{ngcf}\cite{pinsage} will inevitably brings noises to the learned embeddings, degrading the capacity of capturing high-order connectivities.
To be specific, the network layers in GCN fail to distill useful information and features from the aggregated embedding inputs mapped by the one-hot ID features only. Meanwhile, too many parameters in the network layers are prone lead to the issue of overfitting and introduce redundant information into the embedding outputs. As discussed above, the nonlinear network layers in traditional GCN structure is not suitable for recommendation tasks. We elaborate on this in Section \ref{sec:model}.

\subsection{Why Not the Layer Aggregation Mechanism?}
To the best of our knowledge, NGCF\cite{ngcf} is the state-of-the-art GCN-based CF method. In NGCF, layer-aggregation mechanism\cite{layer-aggregation} is applied to concatenate embeddings obtained at each convolution layer as the final embeddings. Despite its effectiveness, we argue that such layer-aggregation mechanism is unnecessary in the CF scenarios when the negative impact of nonlinear network layers is removed. Specifically, the graph convolution can be seen as a linear aggregation process without the network layers. For a target node, the embedding obtained at N-th convolution layer is equivalent to the linear combination of the initial embeddings of all neighbors within N hops, and the concatenation of embeddings obtained at each layer can also be seen as a similar linear combination. As such, using layer-aggregation concatenation mechanism in the CF scenarios is redundant and meaningless. The reason why such layer-aggregation mechanism can work in NGCF is that the information redundancy and noise generated by the network layers can be weakened by the embedding concatenation of each layer. However, in our research we actually find that the nonlinear network structure and the Layer-Aggregation Mechanism limit the model's learning process for high-order connectivities capturing in the CF scenarios. This assumption is detailed in Section \ref{sec:prediction}. In addition, the element-wise product terms $\bm{e}_u \odot \bm{e}_i$ in the aggregation process of NGCF \cite{ngcf} are also redundant to the representations for users and items, we detail this in Section \ref{sec:model}. The above assumptions are verified in Section \ref{sec:rq2}.

\subsection{Our Proposal and Contributions}
In this work, we discussed the limitations of traditional GCN structure for capturing the high-order relations among different entity nodes in recommendation tasks, and propose a new GCN-based CF model, RGCF, where the entities' embeddings are reconstructed with refined graph convolution structure and some strategies are intuitively used to reduce noise and redundancy existed in GCN-based methods. Firstly, a linear weighted average operation is used to instead the complex and nonlinear network layers in the embedding function of the GCN-based methods. Then, we simply use embeddings obtained at last layer as final representations to avoid information overlap, which is caused by embedding concatenation of each convolution layer (Layer-Aggregation Mechanism). Lastly, the element-wise product terms are removed in embedding generating process. In addition, we further improve the model performance by changing the weight of self-loop nodes in the aggregation process on user-item graph. We conduct extensive experiments on the three public datasets, and the results show that RGCF achieves the significant improvement against other state-of-the-art baselines. To be more specific, our model improves over the NGCF w.r.t recall@20 by 17.19\%, 22.18\%, and 40.70\% in Gowalla, Yelp2018, and Amazon-Book respectively.

The main contributions of this work are as follows.
\begin{itemize}
	\color{black}
	\item We analyze and verify the redundancy defect of the GCN-based recommendation methods, and highlight its negative impact on model capability of capturing high-order connectivities.
	\item We present RGCF model to eliminate the representation redundancies inside the GCN-based methods by designing the refined graph convolution structure. In RGCF, the entities' embeddings are more capable of capturing high-order connectivities better than previous methods.   \color{black}
	\item We conduct extensive experiments on three public million-size datasets, empirically demonstrating the state-of-the-art performance of our RGCF.
\end{itemize}

\color{black}The rest of this paper is organized as follows. Section \ref{sec:methology} elaborates our proposed RGCF and discusses the information redundancy. In Section \ref{sec:experiments}, we report the experimental results and analyse the effectiveness and rationality of our proposed RGCF. We give a brief review of related work in Section \ref{sec:relatedwork} and a conclusion of this paper in Section \ref{sec:conclusion}.\color{black}

\section{METHODOLOGY}\label{sec:methology}
\begin{figure*}	
	\centering
	\includegraphics[height=7cm,width=13cm]{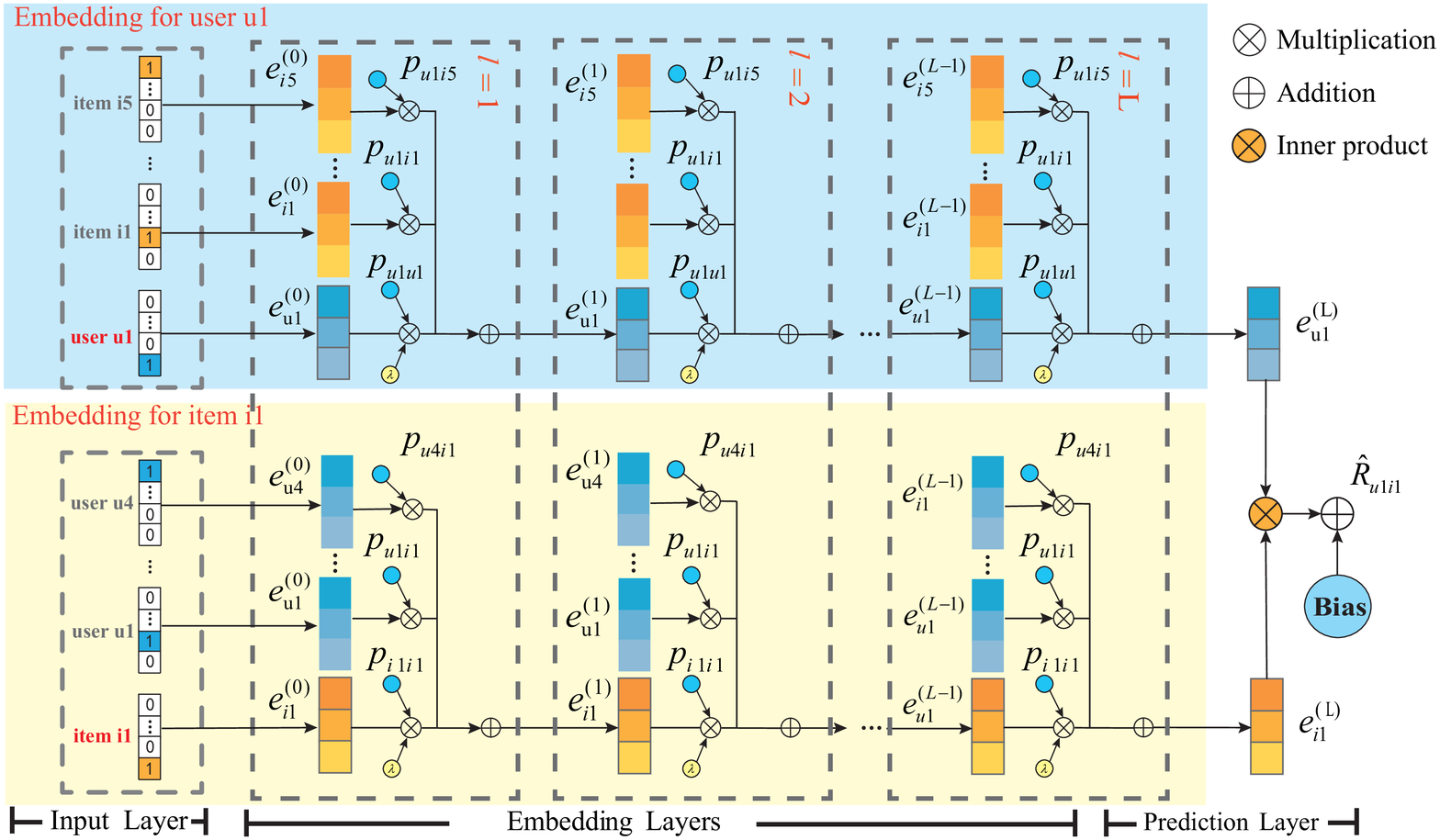}
	\caption{Illustration of our RGCF model which integrates high-order connectivities into the embeddings for user $u1$ and item $i1$ and outputs the matching score for that user-item pair, $\{i1,i2,i3,i4,i5\}$ is the connected item set for user $u1$, and $\{u1,u2,u3,u4\}$ is the connected user set for item $i1$, $p_{uaib}$ equals to $1/\sqrt{\mid N_{ua} \mid \mid N_{ib} \mid} $.}
	\label{fig:model}
\end{figure*}

\color{black}
In this section, we first brief the basic concept of GCN \cite{gcn} and NGCF \cite{ngcf}, and then present our model structure details, as illustrated in Figure \ref{fig:model}. Lastly, we have a discussion about the negative impact of information redundancy on GCN-based method.\color{black}

\subsection{Preliminary}
\color{black}
\noindent\textbf{Graph Convolution Networks}
The core idea of GCN \cite{gcn} is to capture graph structure information by transforming and aggregating information of neighboring nodes. To be specific, GCN includes multiple convolutional layers, in which layer $l+1$ depends on the output of layer $l$. In each layer, the information of the target entity can be aggregated by its neighbor nodes. As such, high-order embeddings can be effectively captured by stacking such multiple convolutional layers. The convolutional operation can be formulated as follows:
\begin{equation}
	\bm{E}^{(l+1)}=\sigma((\bm{D}+\bm{I})^{-0.5}(\bm{A}+\bm{I})(\bm{D}+\bm{I})^{-0.5}\bm{E}^{(l)}\bm{W}^{(l)})
\end{equation}
where $\bm{A} + \bm{I} $ is a $n \times n$ adjacency matrix which a self-loop is added, $n$ is the count of the total nodes, and I is identity matrix, $\bm{D}+ \bm{I}$ denotes the diagonal node degree matrix with elements $\bm{D}_{ii}=1$, $\bm{E}^{(l+1)}$ and  $\bm{E}^{(l)}$ are the $n \times k$ matrix which respectively denote embedding collection obtained at layer $l+1$ and $l$ for all nodes, $k$ is the embedding length.
\color{black}

\noindent\textbf{Neural Graph Collaborative Filtering}
To the best of our knowledge, NGCF \cite{ngcf} is the state-of-the-art GCN-based Collaborative Filtering method. Distinct from standard GCN \cite{gcn}, NGCF integrates the element-wise product of the target nodes and its neighboring nodes into the embedding function, and use the concatenation of the embeddings obtained at each layer as the final representations. The multi-layer-aggregation process for user $u$ can be formulated as follows: 

\begin{align}
	e_u^{(l+1)}=&\sigma( W_1^{(l)}e_u^{(l)} +\sum_{i\in N_u} \frac{1}{\sqrt{|N_u||N_i|}}(W_1^{(l)}e_i^{(l)}+ W_2^{(l)}(e_u^{(l)}\odot e_i^{(l)})))\label{ngcf}
\end{align}

where $W_1^{(l)}$ and $W_2^{(l)}$ are the weight matrices at layer $l$, $e_u^{(l)}$ and $e_u^{(l+1)}$ are the embeddings at layer $l$ and $ l+1$ for current user u, respectively.  $e_i^{(l)}$ is the embedding for item i at layer $l$, $\frac{1}{\sqrt{|N_u||N_i|}}$ is the graph Laplacian norm \cite{gcn} to normalize the embeddings aggregated from previous layer, where $N_u$ and $N_i$ respectively denote u's and i's neighborhood, $\sigma$ is the nonlinear activation function.

\subsection{Model}\label{sec:model}
In this section, we present a detailed description of our RGCF model. As Figure \ref{fig:selfw} shows, the embeddings of users and items are generated separately. User(item) embeddings are generated by propagating information iteratively from the first layer to the last one. In each layer, entity(user or item) is embedded by aggregating the information both from the neighbor nodes and the entity itself.

We use the user embedding construction to detail the aggregation process over the graph (the light blue part of Figure \ref{fig:model}), and the item embedding aggregation is similar. Similar to the convolution operation in GCN, our model outputs the user's embedding by the sum of embeddings of itself and the vectors aggregated by its neighboring nodes. We formulate this iterative embedding process across multiple layer as Algorithm \ref{ag:1}:

\begin{algorithm}[t]
	\caption{Embedding Generating}
	\label{alg:1}
	\hspace*{0.02in} {\bf Input:} Initial embedding $\bm{e_u^{(0)}}$ for node $u$; set of u's neighborhood embeddings $\{\bm{e_i^{(0)}} \mid i \in \bm{N_u}\}$; self-loop weight $\lambda$; and depth of message aggregation layers $L$. \\
	\hspace*{0.02in} {\bf output:} The embedding representation $\bm{e_u^{(L)}}$ obtained at the convolution layer $L$ for node $u$.
	
	\begin{algorithmic}[1] 
		\STATE Let $l=1$.
		\WHILE{$l \neq L$}
		\STATE ${m_{u \leftarrow i}^{(l-1)} = 1/\sqrt{\mid N_u \mid \mid N_i \mid}}  \cdot \sum_{i \in N_u}^{}  e_i^{(l-1)}$.
		\STATE ${m_{u \leftarrow u}^{(l-1)} =1/\sqrt{\mid N_u \mid \mid N_i \mid}}  \cdot e_u^{(l-1)}$
		\STATE ${e_u^{(l)}=m_{u \leftarrow i}^{(l-1)}+\lambda \cdot m_{u \leftarrow u}^{(l-1)}}$
		\STATE $l = l+1$
		
		\ENDWHILE
		\STATE \textbf{return} $\bm{e_u^{(L)}}$
		
	\end{algorithmic}
	\label{ag:1}
\end{algorithm}

Concretely, for target user $u$, we first initialize the embedding of $u$ itself and its neighboring nodes as $e_u^{(0)}$, $\{\bm{e_i^{(0)}} \mid i \in \bm{N_u}\}$ by mapping from ID. Then the embedding representation of user $u$ is iteratively aggregated from Layer 1 to $L$. In Algorithm \ref{ag:1}, Line 3 indicates the message aggregated from the neighboring nodes by using the embeddings from previous layer, and the message that aggregated from the user itself is defined on Line 4.
$m_{u \leftarrow i}^{(l-1)}$ reflects historical interaction information of user $u$, $m_{u \leftarrow u}^{(l-1)}$ can be viewed as the intrinsic properties of the node itself.
We argue that the above two messages have different contributions for generating the final representation for node $u$, thus hyper-parameter $\lambda$ is set to control the weight of the message from $u$ itself. We report the model performance with different parameter settings for $\lambda$ in Section \ref{sec:rq3}.
$1/\sqrt{\mid N_u \mid \mid N_i \mid}$ is the graph Laplacian norm to normalize embeddings aggregated from previous layer, where $N_u$ and $N_i$ respectively denote u's and i's neighborhood. After stacking $L$ such message aggregation operations, we get final representation $e_u^{(L)}$ for node $u$. We can generate the representation for item node $i$ in the similar way.

\noindent\textbf{Matrix Implementation.} In practice, we use sparse matrix multiplications to implement the abovementioned embedding function. The detailed operations can be formulated as follows:
\begin{equation}\label{eq:matrix}
	\bm{E}^{(l)} = (\bm{D}+\lambda \bm{I})^{-0.5}(\bm{A}+\lambda \bm{I})(\bm{D}+\lambda \bm{I})^{-0.5}\bm{E}^{(l-1)}
\end{equation}
where $\bm{A} + \lambda \bm{I} $ is a $(m+n) \times(m+n)$ adjacency matrix in which a weighted self-loop is added, $m$ and $n$ are the number of users and items, and $\bm{I}$ is identity matrix, $\lambda$ is a hyper-parameter to control the weight of self-loop, $\bm{D}+\lambda \bm{I}$ denotes the diagonal node degree matrix with elements $\bm{D}_{ii}=\lambda$, $\bm{E}^{(l)}$ and  $\bm{E}^{(l-1)}$ are the $(m+n) \times k$ matrix which denote embedding collection for all users and items obtained at layer $l$ and $(l-1)$, respectively, and $k$ is the embedding length.
It is worth noting that, distinct from traditional GCN-based methods \cite{ngcf}, the network layers are removed in our embedding generating process since they bring no benefit to model performance.

\subsection{Prediction}\label{sec:prediction}
\color{black}
Distinct from concatenating multiple representations obtained at each convolution layers in NGCF \cite{ngcf}, we use the embeddings obtained at last layer as the final representations, which is same as the standard GCN \cite{gcn}. The key reason is that concatenating representations at different layers may result in the issue of information redundancy. To be specific, the embeddings obtained at layer $l$ actually contain most information comes from previous layers since the aggregation operation at previous layers is a linear operation. \color{black}Thereby, in RGCF, we get the final representations for user $u$ and item $i$ as follows:
\begin{equation}
	\bm{e_u}^*=\bm{e_u}^{(L)},\quad \bm{e_u}^*=\bm{e_u}^{(L)}
\end{equation}   
where $\bm{e_u}^{(L)}$ and $\bm{e_i}^{(L)}$ are the embeddings obtained at last layer $L$ for user $u$ and item $i$ respectively.

Inner product is applied to predict the matching score of a user-item pair $<u, i>$. We formulate the prediction function as follows:
\begin{equation}
	\hat{\bm{r}}_{ui}={\bm{e_u}^{*}}^T \bm{e_i}^{*}+\bm{b_u} + \bm{b_i}
\end{equation}

where $\hat{r}_{ui}$ is a predicted preference score for $u$ towards the target item $i$, $\bm{e_u}^{*}$ and $\bm{e_i}^{*}$ are the final representations for user $u$ and item $i$, $\bm{b_u}$ and $\bm{b_i}$ denote the bias for $u$ and $i$, respectively. 
\color{black}Note that setting bias terms can help distinguish nodes with different popularity since that the nodes with a large number of interactions can learn a larger bias than the nodes with a small number of interactions. That is to say, the value of the bias depends on the popularity of the nodes. This term can alleviate the negative impact of oversmoothing and improve model performance especially for top-N ranking task in recommendation.\color{black}

\subsection{Training}\label{sec:training}
\noindent\textbf{Loss Function.}
We use Bayesian Personalized Ranking(BPR) loss \cite{bprloss} to optimize the parameters for our model. The basic assumption for BPR loss is that the observed interactions can reflect stronger preference than unobserved ones, that is to say, the predicting score for an observed user-item pair should be higher than that of the unobserved one. The loss function for our model is formulated as follows:
\begin{equation}
	loss=\sum_{(u,i,j)\in O}^{}-\ln \sigma (\hat{y}_{ui}-\hat{y}_{uj})+\alpha {\mid\mid E\mid\mid}_2^2 + \beta {\mid\mid B\mid\mid}_2^2
\end{equation}

where $O=\{(u,i,j) \mid i\in N_u, j \notin N_u \}$ is the training data, $N_u$ denotes the observed item set for user $u$, $\sigma(\cdot)$ is the sigmoid function; we apply $L_2$ regularization on $E$ and $B$ parameterized by $\alpha$ and $\beta$ respectively, $E$ and $B$ are the final embeddings obtained at last layer and the biases for all users and items respectively.

\noindent\textbf{Optimizer.} Mini-batch Adam optimizer \cite{adam} is applied to optimize our model and update the model parameters. Note that the parameters that need to be updated are the embeddings mapped from ID and the biases for all users and items, which is almost equals to that of BiasSVD \cite{biassvd}.

\subsection{Discussion on Information Redundancy}
\label{sec:redundacny}

\noindent\textbf{Why network layer is redundant?}
Distinct from traditional GCN-based methods, the nonlinear network layers are removed in our RGCF since they bring no benefit to model performance. Although the network layers can find hidden patterns from complex input embeddings which usually contain rich side information, the expressiveness of the embeddings will be limited if the inputs do not have complex patterns ($aka.$ embeddings mapping from ID mapping). Meanwhile, the overfitting problem caused by too many parameters of network layers cannot be completely eliminated even if dropout technology is applied. 

\noindent\textbf{Why layer-aggregation mechanism is redundant?}\label{sec:redundancy}
Because the embedding aggregation at each layers is a linear transformation, the embeddings obtained at layer $l$ already contain the information inside the embeddings of its previous layers. 
As such, embeddings concatenation of each layer is equivalent to multifoldly consider the contribution of low-order interactions, where the contribution of high-order interactions are relatively weakened consequently. This kind of analysis nicely supports our argument that redundancies exist in traditional GCN-based recommendation methods lead to poor of high-order connectivies capturing capacity. We use the following simplified formula which ignores the influence of graph Laplacian norm to justify this assumption. 
\begin{align}
	\bm{E}^{(2)} =& (\bm{A}+\bm{I})(\bm{A}+\bm{I})\bm{E}^{(0)}\notag\\ 
	=& \bm{A}(\bm{A}+\bm{I})\bm{E}^{(0)} + (\bm{A}+\bm{I})\bm{E}^{(0)}\notag\\ 
	=& \bm{A}(\bm{A}+\bm{I})\bm{E}^{(0)} + \bm{E}^{(1)}
\end{align}
where $\bm{E}^{(1)}$ and $\bm{E}^{(2)}$ denote embedding matrices obtained at first layer and second layer. We can see that $\bm{E}^{(2)}$ contains $\bm{E}^{(1)}$. \color{black}In this way, concatenating embedding of each layer is unnecessary when network layers are removed in RGCF. It is worth mentioning that the concatenation operation in NGCF can be effective. 
This is because that in NGCF the defective embeddings impaired by non-linear network layer may be remedied by the concatenation operation to some extent
we conduct some experimental comparison in Section \ref{sec:rq2} to verify this assumption.\color{black}

\noindent\textbf{Why product term is redundant?}
In NGCF, the product term $\bm{e}_u \odot \bm{e}_i$ in equation \ref{ngcf} magnify the preference score of the user-item pair, which can increase the affinity of the interacted nodes and help speed up the model convergence. In addition, this term can weaken the negative impact of the information redundancy and noise generated by nonlinear graph convolution, which is similar to the abovementioned concatenation operation. In fact, such product term is also redundant while the interaction function is inner product. To be specific, the result of the inner product of $\bm{e}_u$ and $\bm{e}_i$ can reconstruct the information of the product term $\bm{e}_u \odot \bm{e}_i$. We further verify this assumption in Section \ref{sec:rq2}.

\section{EXPERIMENTS}\label{sec:experiments}
In this section, we conduct experiments on three public datasets to evaluate the performance for our proposed model. We aim to answer the following research questions:
\begin{itemize}
	\item \textbf{RQ1}: How does our proposed RGCF perform compared to other state-of-the-art CF models?
	\item \textbf{RQ2}: Whether the refined graph convolution structure is helpful for capturing high-order connectivities and further improving the model performance?
	\item \textbf{RQ3}: How do the key hyper-parameter settings affect the performance of our proposed RGCF?
\end{itemize}

\subsection{Experimental Settings}
\noindent\textbf{Dataset Description.} We conduct experiments on three datasets: Gowalla \cite{gowalla}, Yelp2018, and Amazon-book, which are the same as that used in NGCF \cite{ngcf}. We show the statistics of the three datasets in Table \ref{tab:statistics}. To ensure the quality of the dataset, 10-core setting is applied to retain the users and items with at least ten interactions. For each dataset, we sample $80\%$ of historical interactions for each user as the training set, and treat the remaining $20\%$ as the test set, meanwhile, we resample $10\%$ of historical interactions from the training data as the validation set to tune the hyper-parameters.

\renewcommand{\arraystretch}{1.5}
\begin{table}
	\centering
	\fontsize{10}{10}\selectfont
	\caption{Statistics of the datasets.}
	\label{tab:statistics}
	\begin{tabular}{c|c|c|c|c}
		\hline
		\bf{Dataset}&\bf{User \#}&\bf{Item \#}&\bf{Interaction \#}&\bf{Density}\cr
		\hline
		\hline
		\bf{Gowalla}&29,858&40,981&1,027,370&0.00084\cr
		\hline
		\bf{Yelp2018}&31,668&38,048&1,561,406&0.00130\cr
		\hline
		\bf{Amazon-Book}&52,643&91,599&2,984,108&0.00062\cr
		\hline
	\end{tabular}
\end{table}

\noindent\textbf{Evaluation Metrics.} We select two widely-used evaluation protocols \cite{ngcf}: recall$@K$ and ndcg$@K$ to evaluate the model performance. Specifically, we compute the average recall$@K$ and ndcg$@K$ for each user in the test set.

\noindent\textbf{Baselines.} We compare our proposed method with the following baselines:
\begin{itemize}
	\item \textbf{MF} \cite{mf}: This is a matrix factorization method with Bayesian Personalized Ranking(BPR) loss, which is widely used for recommendation baseline.
	
	\item \color{black}\textbf{SVD++} \cite{svd++}: It is a variant of MF, which uses the user's historical interactions to model the user's preferences. It can also be regarded as a one-layer GCN, and it only passes messages for user embeddings. For fairly comparison, we use Bayesian Personalized Ranking(BPR) loss to optimize svd++.\color{black}
	
	\item \textbf{NeuMF} \cite{ncf}: It is a state-of-the-art neural collaborative filtering method which uses nonlinear neural networks as interaction function.
	
	\item \textbf{HOP-Rec} \cite{hoprec}: It is a state-of-the-art graph-based method, which uses random walk to enrich the interaction data between users and their multi-hop connected items.
	
	\item \textbf{GC-MC} \cite{gcmc}: This model adopts GCN technique which just contains one layer convolution operation to generate the users and items representations. 
	
	\item \textbf{NGCF} \cite{ngcf}: It is a state-of-the-art GCN-based method, which combines embeddings obtained at different GCN layer as the final users and items representations.
	
\end{itemize}

\noindent\textbf{Parameter Settings.} To make a fair comparison, we set the embedding size as 64 for all models. We apply a grid search strategy to tune the following hyper-parameters: the learning rate is searched in $\{$0.001,0.0005,0.0001,0.00005$\}$,  the coefficient of $L_2$ normalization is searched in $\{ 1,0.1,...,10^{-6},10^{-7} \}$, and the weight of self-loop is searched in $\{$ 0.0,0.3,0.5,0.7,1.0,1.2,1.5,1.7,2.0 $\}$. In addition, We use early stopping strategy to prevent overfitting. Our experiment results show that the optimal learning rate is $0.001$ and the optimal coefficient of $L_2$ normalization is $10^{-3}$ for Gowalla, $10^{-4}$ for Yelp2018, and $10^{-5}$ for Amazon-book respectively.

\renewcommand{\arraystretch}{1.5}
\begin{table*}[tp]
	
	\centering   
	\fontsize{10}{10}\selectfont
	\caption{Overall performance comparison w.r.t. recall@20 and ndcg@20 on Gowalla, Yelp2018, and Amazon-Book.}
	\label{tab:overall}
	\begin{tabular}{|c|c|c|c|c|c|c|}
		\hline
		\multirow{2}{*}{\bf{Method}}&
		\multicolumn{2}{c|}{\bf{Gowalla}}&\multicolumn{2}{c|}{\bf{Yelp2018}}&\multicolumn{2}{c|}{\bf{Amazon-Book}}\cr\cline{2-7}
		&\bf{recall}&\bf{ndcg}&\bf{recall}&\bf{ndcg}&\bf{recall}&\bf{ndcg}\cr
		\hline
		\hline
		{MF}&0.1291&0.1878&0.0433&0.0864&0.0250&0.0518\cr
		{SVD++}&0.1439&0.2198&0.0507&0.0975&0.0332&0.0607\cr
		{NeuMF}&0.1326&0.1985&0.0449&0.0886&0.0253&0.0535\cr
		\hline
		{HOP-Rec}&0.1399&0.2128&0.0524&0.0989&0.0309&0.0606\cr\hline
		{GC-MC}&0.1395&0.1960&0.0462&0.0922&0.0288&0.0551\cr
		{NGCF}&0.1547&0.2237&0.0559&0.1037&0.0344&0.0630\cr
		\hline
		\textbf{Ours}&{\textbf{0.1813}}&{\textbf{0.2457} }&{\textbf{ 0.0683}}&{\textbf{0.1212} }&{\textbf{0.0484} }&{ \textbf{0.0840}}\cr
		\hline
		\hline
		\%Improv.&17.19\%&9.83\%&22.18\%&16.87\%&40.70\%&33.33\%\cr\hline
		
	\end{tabular}
	
\end{table*}
\subsection{Performance Comparison (RQ1)}
\label{sec:rq1}
We compare the performance of all the methods in this section. Table \ref{tab:overall} reports the performance of recall@20 and ndcg@20 for all compared methods. We have the following findings:  
\begin{itemize}
	\item MF achieves poor performance in three datasets, indicating that simple inner product is insufficient to capture complex connectivities between users and items. NeuMF outperforms MF across all datasets, which verifies the effectiveness of applying neural networks to distill the nonlinear relations between users and items.
	
	\item Compared to MF and NeuMF, the performance of GC-MC demonstrates that integrating the first-order connectivities into the embedding process is helpful for improving the expressiveness of the embeddings.
	
	\item HOP-Rec generally outperforms GC-MC across all cases. The key reason is that HOP-Rec exploits the high-order neighbors to enrich the training data while GC-MC considers the first-order neighbors only. 
	
	\item \color{black}The performance of SVD ++ is significantly better than that of GC-MC, which also verifies that using a nonlinear network layer to process id embeddings will add information redundancy and noise to the representations, thereby degrading the model performance.  \color{black}
	
	\item NGCF consistently outperforms HOP-Rec, which demonstrates that explicitly integrating high-order connectivities into the embedding process is more efficient than exploiting high-order interactions to enrich the training data. \color{black}Meanwhile, the performance of NGCF is slightly higher than that of SVD ++. The main reason is that NGCF integrates high-order interactions into the embeddings of users and items, while SVD ++ only integrates first-order interactions into user embeddings.\color{black}
	
	\item Our RGCF yields the best performance on all the datasets compared to all the baselines. Specifically, RGCF improves over the strongest baselines w.r.t. recall@20 by $17.19\%$, $21.46\%$, and $34.88\%$ in Gowalla, Yelp2018, and Amazon-Book, respectively. The significant improvements across all cases verify that our refined graph convolution structure and other strategies to reduce noise and information redundancy is rational and effective.
\end{itemize}

\subsection{Is Refined Graph Convolution Structure Effective? (RQ2)}\label{sec:rq2}
In this section, we first verify that the three components in GCN-based methods introduced in section \ref{sec:redundacny} are redundant. And then, we set the experimental comparison $w.r.t.$ different number of convolution layers to verify whether our proposed RGCF can enhance the capacity of high-order connectivities capturing.

\begin{table*}  
	\centering
	\fontsize{10}{10}\selectfont
	\caption{Performance of RGCF variants with different information redundancy.}
	\label{tab:redundancy}
	\begin{tabular}{|c|c|c|c|c|c|c|}
		\hline
		\multirow{2}{*}{\bf{Method}}&
		\multicolumn{2}{c|}{\bf{Gowalla}}&\multicolumn{2}{c|}{\bf{Yelp2018}}&\multicolumn{2}{c|}{\bf{Amazon-Book}}\cr\cline{2-7}
		&\bf{recall}&\bf{ndcg}&\bf{recall}&\bf{ndcg}&\bf{recall}&\bf{ndcg}\cr
		\hline
		\hline
		{RGCF+np}&0.0584&0.0777&0.0216&0.0465&0.0164&0.0355\cr
		{RGCF+n}&0.0608&0.0825&0.0238&0.0515&0.0166&0.0354\cr
		{RGCF+npc}&0.1547&0.2237&0.0559&0.1037&0.0344&0.0630\cr
		{RGCF+nc}&0.1616&0.2361&0.0562&0.1041&0.0359&0.0646\cr
		\hline
		{RGCF+pc}&0.1579&0.2314&0.0584&0.1073&0.0366&0.0675\cr
		
		{RGCF+p}&{0.1665}&{0.2392}&{0.06246}&{0.11392}&{0.03802}&{0.07092}\cr
		
		{RGCF+c}&{0.1680}&{0.2334}&{0.0585}&{0.1072}&{0.0373}&{0.0689}\cr
		\hline
		{RGCF}&{ 0.1813}&{ 0.2457}&{ 0.0683}&{ 0.1212}&{ 0.0484}&{ 0.0840}\cr
		\hline
	\end{tabular}
\end{table*}

\subsubsection{\textbf{Impact of information redundancy.}}
We have analyzed the redundancy issues of some state-of-the-art GCN models in Section \ref{sec:redundacny}, namely (1) non-linear network layers redundancy, (2) embedding concatenation redundancy, and (3) element-wise product redundancy.
\color{black}
For the sake of presentation, we divide the experiment into two parts: Part A with non-linear network layers, and Part B without non-linear network layers. 

For experiments with network layers(Part A), we have the following derived model from RGCF:
\begin{itemize}
	\item \textbf{RGCF+n} denotes the variant model in which only the non-linear network layers redundancy is reserved.
	\item \textbf{RGCF+np} denotes the variant model with the redundancies of  non-linear network layers and product terms, means that embedding concatenation redundancy is removed in NGCF.
	\item \textbf{RGCF+nc} denotes the variant model with the redundancies of non-linear network layers and the embedding concatenation, means that product term redundancy is removed in NGCF.
	\item \textbf{RGCF+npc} denotes the variant model which contains the above three redundancies, which is equivalent to NGCF.
\end{itemize} 
For experiments without non-linear network layers(Part B), we have the following derived model from RGCF similarly: 
\begin{itemize}
	\item \textbf{RGCF+c} denotes the variant model in which only the embedding concatenation redundancy is reserved.
	\item \textbf{RGCF+p} denotes the variant model in which only the product terms redundancy is reserved.
	\item \textbf{RGCF+pc} denotes the variant model with the redundancies of the embedding concatenation and product terms.
	\item \textbf{RGCF} indicates that the three redundancies are all removed.
\end{itemize}
		

\begin{figure*}[htbp]
	\centering
	\subfigure[Gowalla-recall.]{
		\centering
		\includegraphics[width=1.7in]{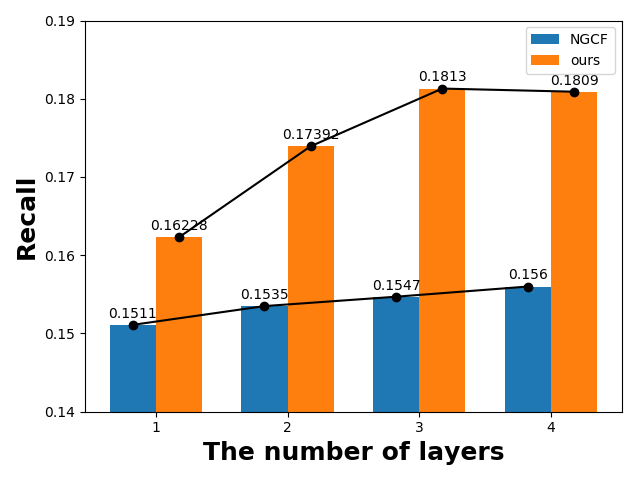}
	}%
	\subfigure[yelp2018-recall.]{
		
		\centering
		\includegraphics[width=1.7in]{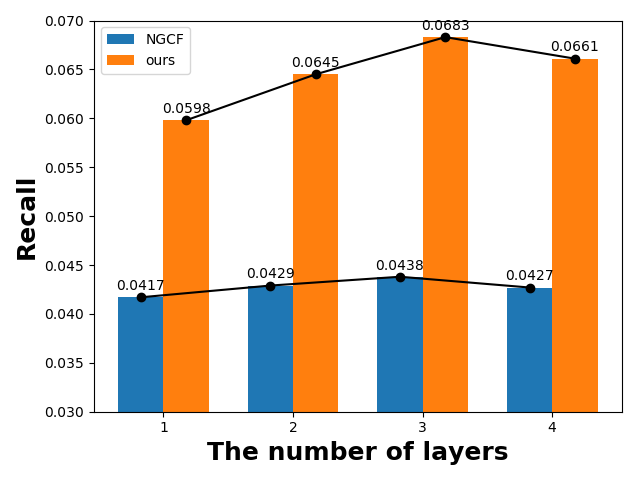}
	}%
	\subfigure[amazon-book-recall.]{
		
		\centering
		\includegraphics[width=1.7in]{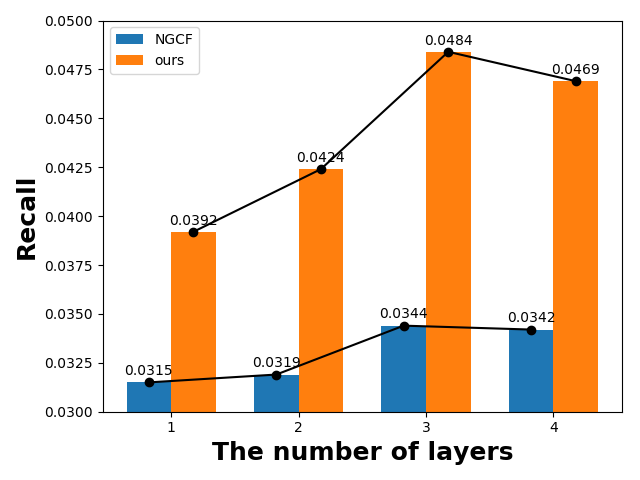}
	}
	\subfigure[Gowalla-ndcg.]{
		\centering
		\includegraphics[width=1.7in]{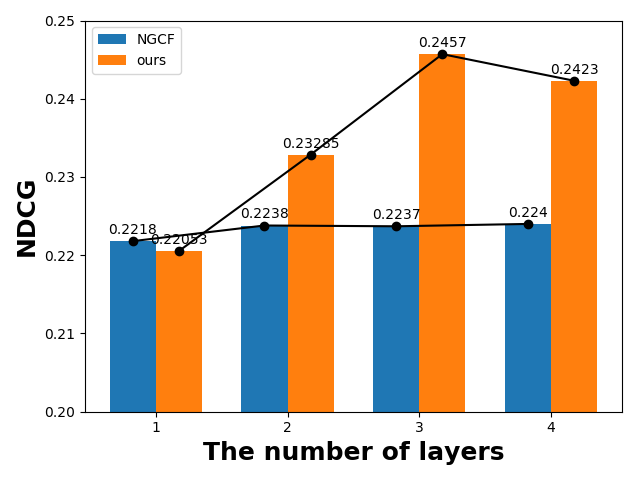}
	}%
	\subfigure[yelp2018-ndcg.]{
		\centering
		\includegraphics[width=1.7in]{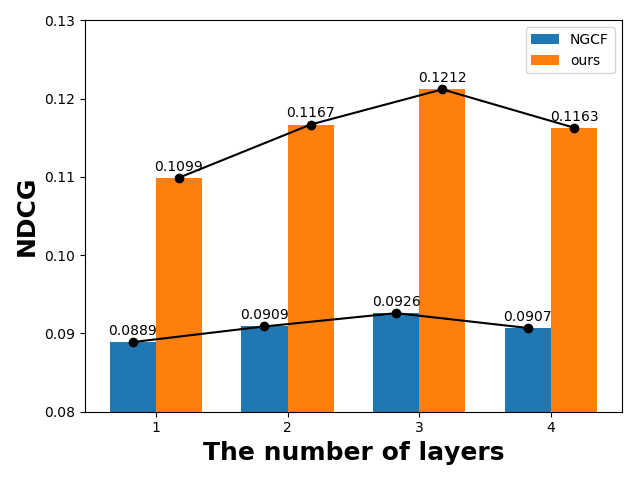}
	}%
	\subfigure[amazon-book-ndcg.]{
		\centering
		\includegraphics[width=1.7in]{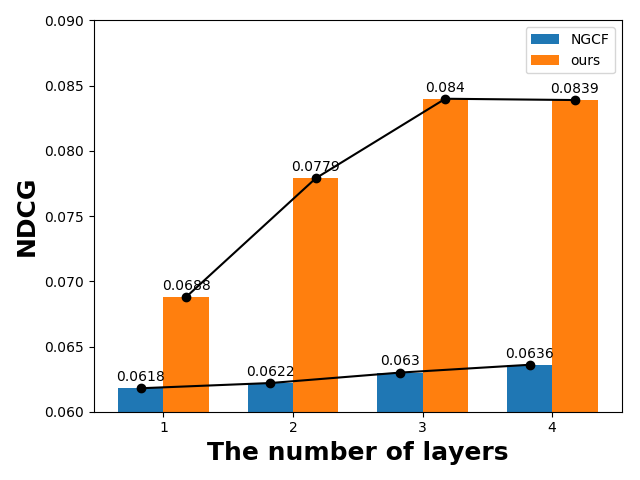}
	}%
	\centering
	\caption{ Performance of NGCF and RGCF with different number of convolution layers $L$ w.r.t. $recall@20$ and $ndcg@20$ on Yelp2018 and Gowalla.}
	\label{fig:rq3-2}
	\centering
	
\end{figure*}

Table \ref{tab:redundancy} reports the experimental results. We have the following findings: 

\begin{itemize}
	\item RGCF+n slightly outperforms RGCF+np, RGCF+nc slightly outperforms RGCF+npc, and RGCF+c outperforms RGCF+pc, which all indicate that element-wise product term is redundant and brings no benefit to model performance. 
	
	\item Compared to RGCF+np and RGCF+n, RGCF+npc and RGCF+nc achieve significant improvements. On the contrary, RGCF+p outperforms RGCF+pc. This result verifies the abovementioned assumption in Section \ref{sec:redundancy}. Specifically, embedding concatenation can partially remedy the impairing effect on embedding quality caused by non-linear network layers. However, RGCF+p outperforms RGCF+npc, which means that removing both redundancies at the same time can further improve the model performance.
	
	\item Compared to RGCF+npc, RGCF+pc achieves the better performance. This result demonstrates that the network layers in GCN fail to extract the useful features from the inputs mapped by ID and further limit the model performance.
	
	\item RGCF+c and RGCF+p slightly outperforms RGCF+pc across all the cases, which verifies that the product terms in NGCF and concatenation operation easily lead to the issue of information redundancy and removing these terms from embedding function can facilitate the recommendation task.
	
	\item RGCF consistently achieves the best performance. This result demonstrates that our refined graph convolution structure which eliminates the above three redundancy issues can greatly enhance the learning process of high-order connectivities and further improve the recommendation cases.
\end{itemize}

\subsubsection{\textbf{Effect of convolution layer numbers}}
To illustrate the impact for RGCF $w.r.t.$ the number of convolution layers $L$, we demonstrate the experimental result $w.r.t.$ Recall@20 and NDCG@20 on Gowalla and Yelp2018 with different $L$ in Figure \ref{fig:rq3-2}. Jointly analysing the Figure \ref{fig:rq3-2}, we have the following observations:

\begin{itemize}
	\item The performance of NGCF and RGCF $w.r.t.$ Recall@20 and NDCG@20 is improved significantly with the increasing of the depth of layers in most cases. Such result demonstrates that high-order interaction is essential for modeling user preference.
	
	\item As the depth of layers increases, the performance of NGCF improves slightly, while RGCF has impressive improvement across all the cases. This is because that our RGCF model can benefit much more from the growth of the layers depth than NGCF, verifying again that the refined structure in RGCF is capable of capturing the high-order connectivities in the user-item interaction graph.
	
	\item When the depth of layers increases to four, both the model performance of both RGCF and NGCF slightly decrease due to overfitting. such result shows that conducting three graph convolution layers is sufficient to model expressive embeddings for users and items.
\end{itemize}

\subsection{Study of hyper-parameters (RQ3)}  
\label{sec:rq3}
\color{black}
In this study, we investigate the impact of different self-loop weight $\lambda$ and $L_2$ regularization coefficient on the performance of our proposed model. 
\color{black}

\subsubsection{\textbf{Effect of self-loop weight}}
\begin{figure*}[t]
		
		\centering
		\subfigure[Gowalla.]{
			
			\centering
			\includegraphics[width=1.7in]{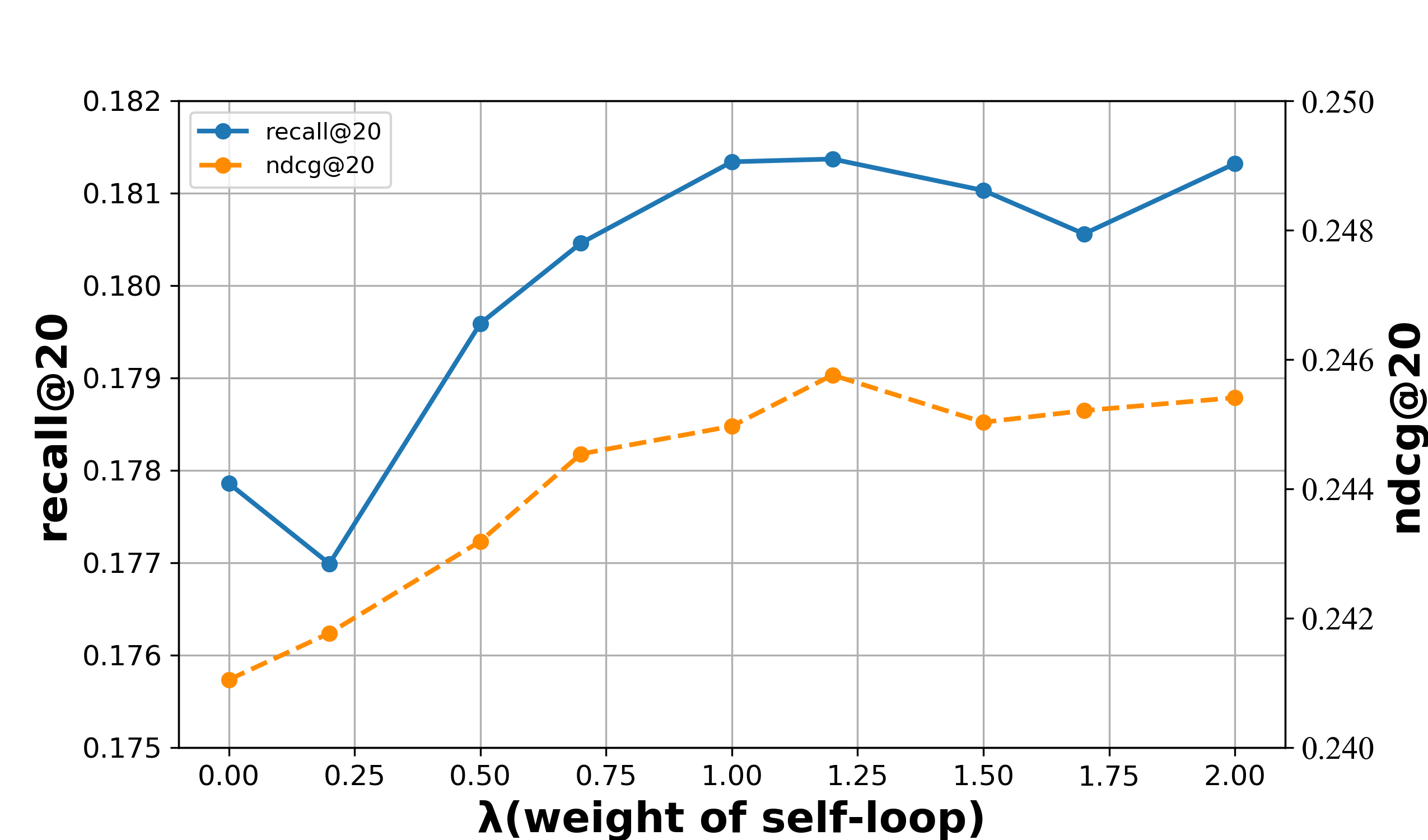}
		}%
		\subfigure[Yelp2018.]{
			
			\centering
			\includegraphics[width=1.7in]{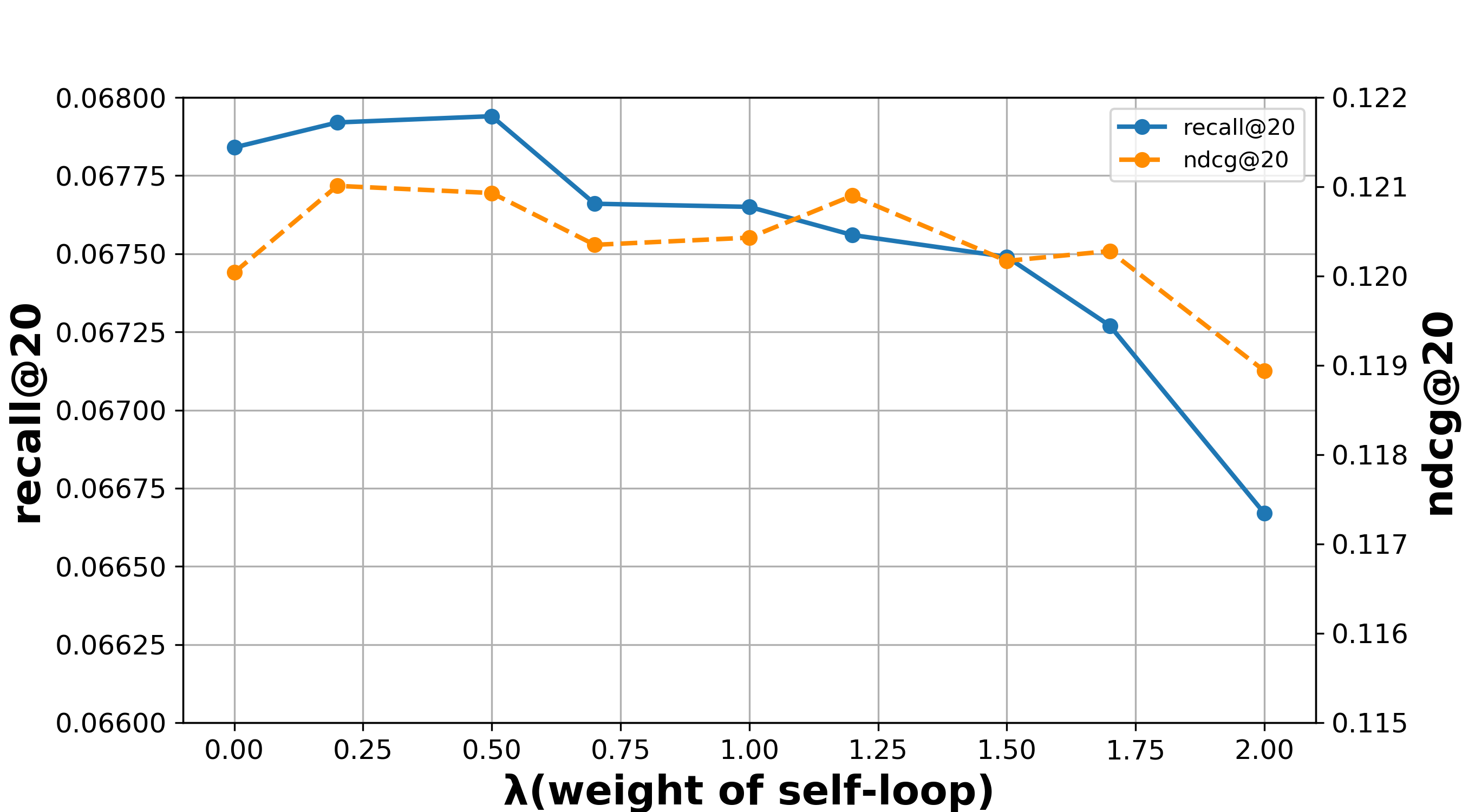}
		}%
		\subfigure[Amazon-Book.]{
			
			\centering
			\includegraphics[width=1.7in]{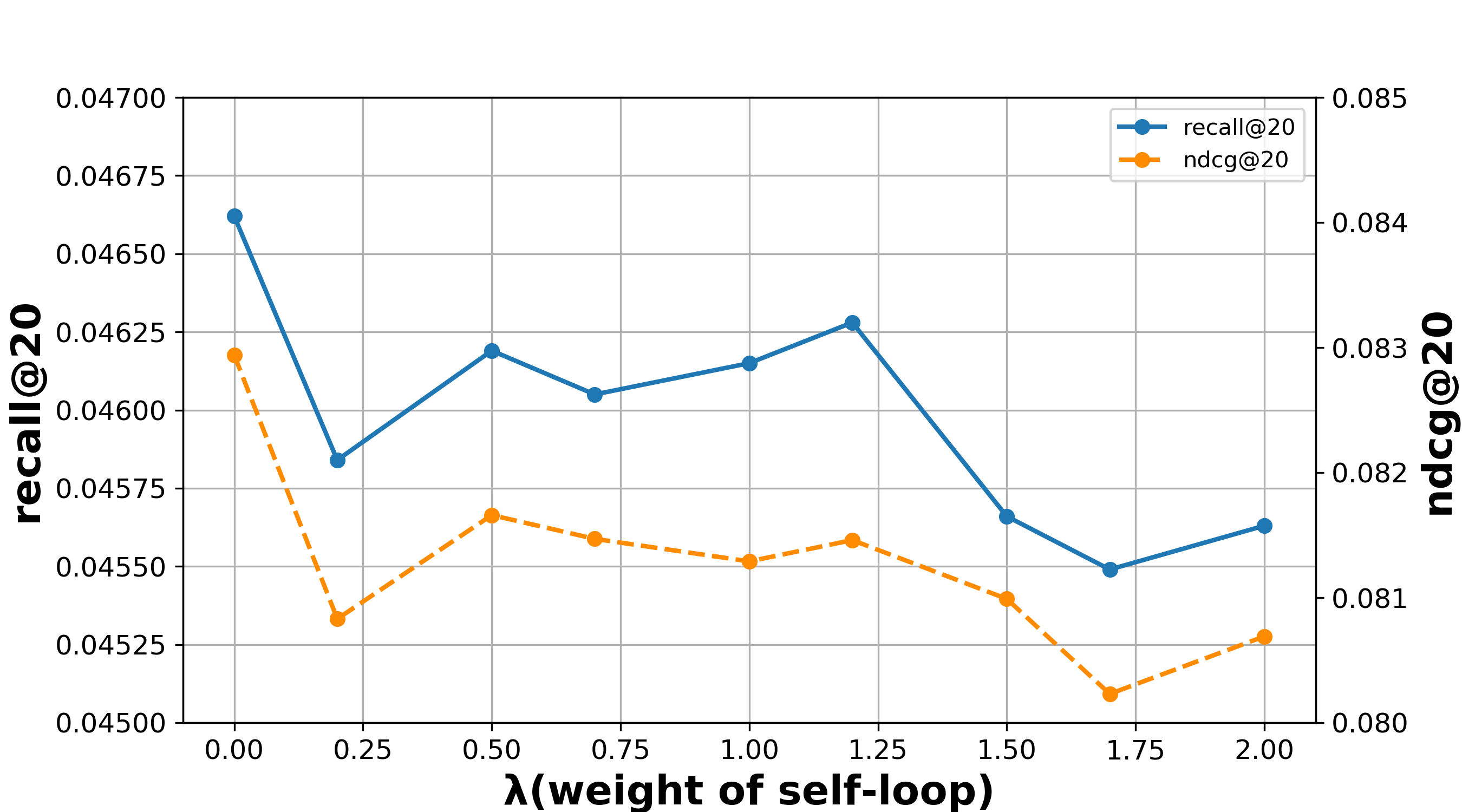}
		}%
		\centering
		\caption{ Performance of RGCF with different self-loop weights $\lambda$ w.r.t. $recall@20$ and $ndcg@20$ on Yelp2018, Gowalla, and Amazon-Book.}
		\label{fig:selfw}
	\end{figure*}

To investigate how self-loop weight affects the model performance. We search the $\lambda$ in the range of $\{ 0.0,0.2,0.5,0.7,1.0,1.2,1.5,1.7,2.0 \}$. Figure \ref{fig:selfw} plots the effect of self-loop weight $\lambda$ w.r.t. $recall@20$ and $recall@20$ on the three datasets.
Specifically, our RGCF achieves the best performance when $\lambda=1.2$ for Gowalla, $\lambda=0.5$ for Yelp2018, and $\lambda=0.0$ for Amazon-Book, respectively. Such experimental result shows that the importance of self-loop is different on different datasets. Therefore finding a appropriate value of self-loop weight can be an effective strategy to further improve the recommendation task.

\subsubsection{\textbf{Effect of $L_2$ regularization coefficient}}
\begin{figure*}[t]
		\centering
		\subfigure[Gowalla.]{
			\centering
			\includegraphics[width=1.7in]{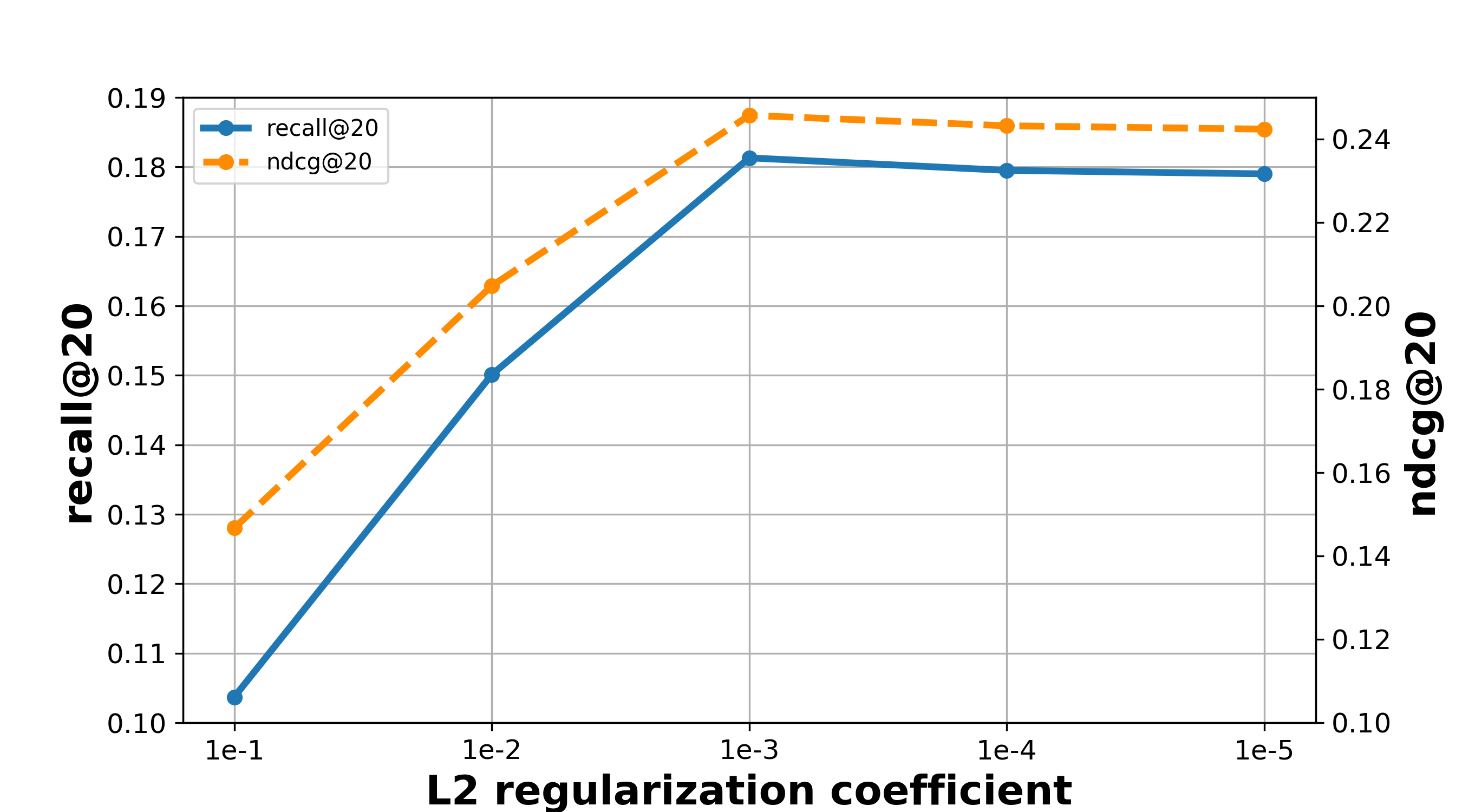}
		}%
		\subfigure[Yelp2018.]{
			\centering
			\includegraphics[width=1.7in]{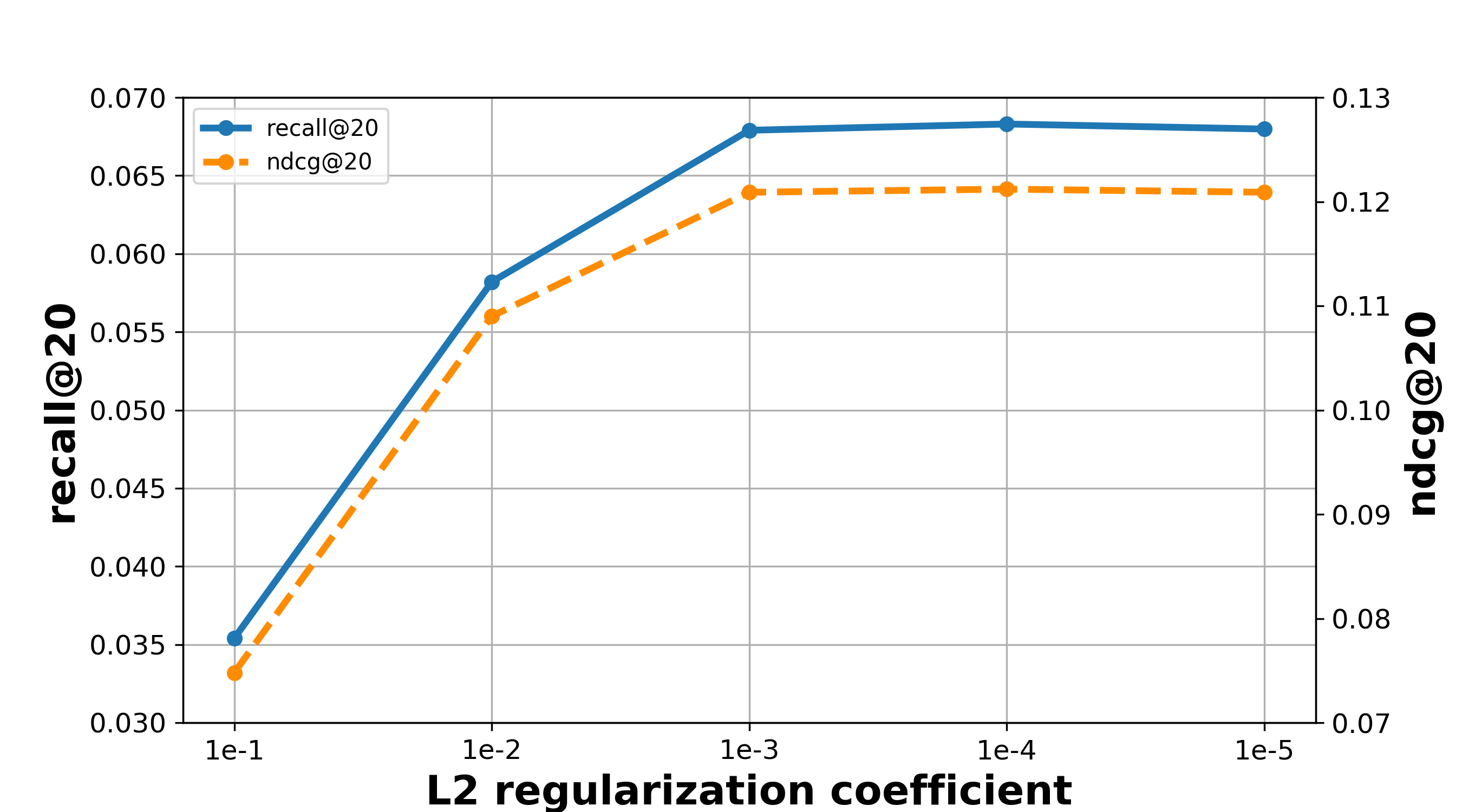}
		}%
		\subfigure[Gowalla.]{
			\centering
			\includegraphics[width=1.7in]{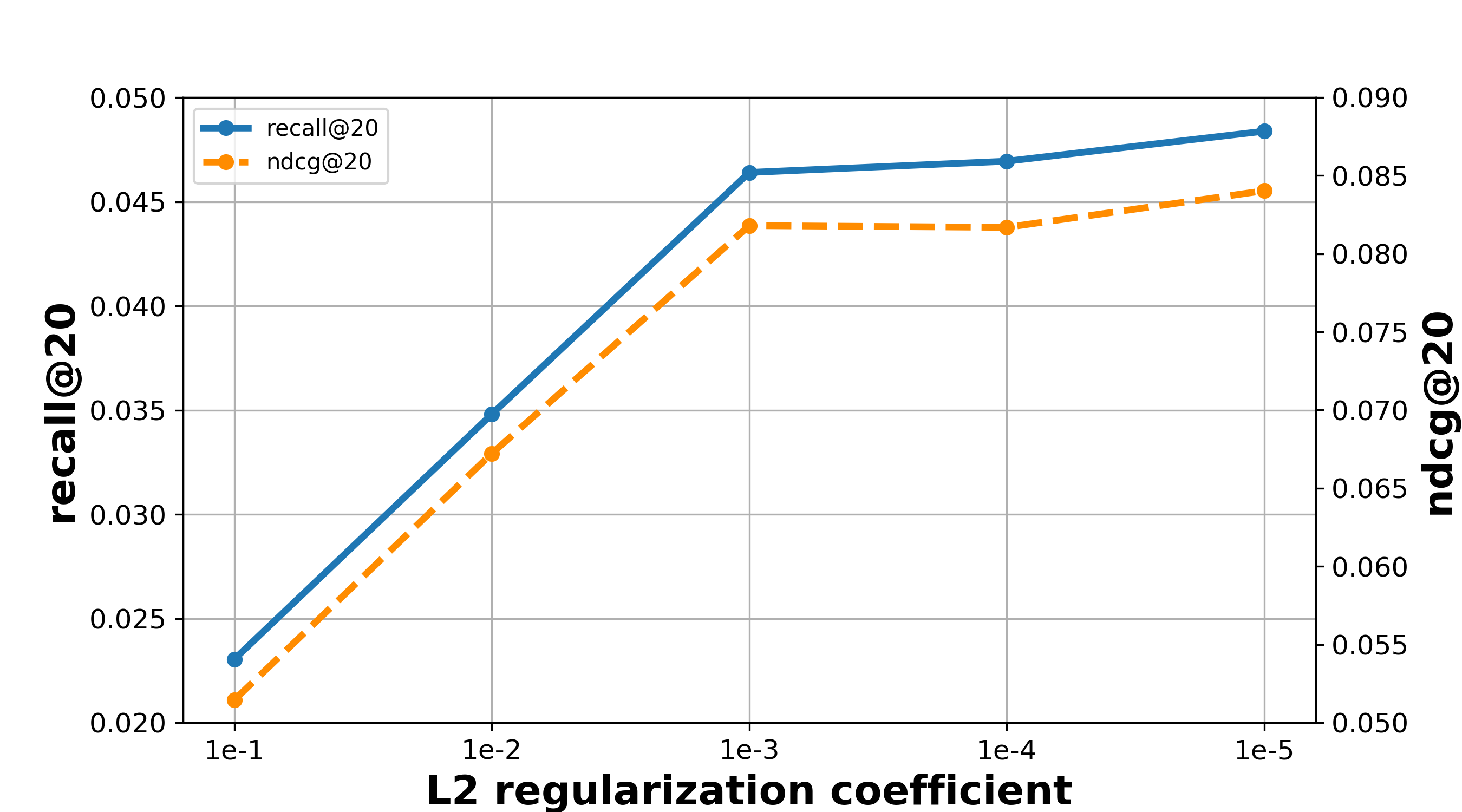}
		}%
		\caption{ Performance of NGCF and RGCF with different $L_2$ regularization coefficient $\alpha$ w.r.t. $recall@20$ and $ndcg@20$ on Yelp2018, Gowalla, and Amazon-book.}
		\label{fig:rq3-3}
		\centering
		
	\end{figure*}

Figure \ref{fig:rq3-3} show the test performance $w.r.t.$ recall@20 and NDCG@20 of RGCF with regarding to different $L_2$ regularization coefficient settings on three datasets. We tune the $L_2$ regularization coefficient $\alpha$ in the range of $\{1e-1,1e-2,1e-3,1e-4,1e-5\}$. From the experimental results, We found that RGCF achieves the best performance when $\alpha=0.001$ for Gowalla, $\alpha=0.0001$ for yelp2018, and $\alpha=0.00001$ for Amazon-book respectively. 
\color{black}


\section{RELATED WORK}\label{sec:relatedwork}
This section introduces factorization-based CF methods and GCN-based CF methods, which are most related to our work.

\subsection{Factorization-based CF methods}
The core idea of the factorization-based methods is to parameterize all users and items and use the product of the user matrix and the item matrix to reconstruct the interaction matrix. For example, Matrix Factorization (MF) \cite{mf} obtains vector representations of users and items by mapping their IDs. In order to improve the expressiveness of user embeddings, SVD++ integrates historically interacted item embeddings into user embeddings \cite{svd++}. Meanwhile, many works believe that some auxiliary properties which are related to users and items, such as age, gender, occupation, price and multimedia feature \cite{visual1}\cite{visual2}, are relevant to user preferences, and integrate such properties into the embeddings to improve the model performance. Despite the effectiveness of the abovementioned methods, these methods ignore the importance of modeling high-order connectivities. Some works can capture such high-order connectivities. For example, HOSLIM \cite{hoslim} encodes high-order interactions into the embeddings but the time complexity is too high to handle the million-size dataset. DICF \cite{dicf} and NCF \cite{ncf} apply the nonlinear neural netowrks as interaction function to capture high-order interactions. HOP-Rec \cite{hoprec} is a fusion algorithm of graph method and matrix factorization method, which uses the random-walk to find higher-order neighboring nodes as a positive sample of the target node, achieving convincing results. However, HOP-Rec only uses high-order interactions to enrich the training data, the embedding representations of users and items lack explicit encoding of higher-order connectivities.

\subsection{GCN-based CF methods}
The GCN-based methods \cite{gcn}\cite{sgc}\cite{graphsage} are capable of capturing high-order interaction connectivities between graph nodes, which is integrated into the node embedding representations. In recent years, many works have applied GCN techniques to the research field of recommendation system. GC-MC \cite{gcmc} uses GCN to construct an encoder to aggregate the information of first-order neighbors into the embedding representations of the target nodes. Compared with GC-MC, PinSage\cite{pinsage} extends the message aggregation function to higher-order cases and achieves the better model performance. The Section 4.4.1 in NGCF \cite{ngcf} has proven that high-order neighborhood information aggregation can improve the expressiveness of the embeddings. NGCF is a new work that combines GCN and MF to integrate high-order connectivities into the users and items embedding representations and predict the preference score with the inner product of them. 

\color{black}
Despite their effectiveness, we theoretically and empirically find that these methods suffer from some redundancy problems discussed in section \ref{sec:redundacny} and the capability of capturing high-order connectivities is suboptimal. We design a refined graph convolution structure to avoid these information redundancy problems and achieve significant performance improvement shown in Table \ref{tab:overall}.
\color{black}

\section{Conclusion}\label{sec:conclusion}
In this work, we highlight that refined graph convolution in the embedding generating process and other strategies to reduce information redundancy are critical important to enhance the model capability of capturing high-order connectivities, and further improve the expressiveness of the embeddings for users and items. We present a new GCN-based CF model, RGCF, which alleviates the negative impact caused by information redundancy and achieves significant improvements against other state-of-the-art recommendation models. Experimental results and further analysis demonstrate the effectiveness and rationality of our proposed RGCF. 
\color{black}

In future work, we wish to further improve the RGCF performance using the attention mechanism \cite{attention1}\cite{attention2} to precisely assign the weight for neighboring nodes. Meanwhile, we are interested in integrating the causal inference \cite{relation2} and knowledge graph \cite{kg1}\cite{kg2} into our RGCF to improve the interpretability in recommendation.

 \bibliographystyle{ACM-Reference-Format}
 \bibliography{sample-base}

\end{document}